\documentclass{ecai}
\usepackage{hyperref}
\usepackage[small]{caption}
\usepackage{latexsym}
\usepackage{amssymb}
\usepackage{amsmath}
\usepackage{amsthm}
\usepackage{booktabs}
\usepackage{enumitem}
\usepackage{graphicx}
\usepackage{color}
\usepackage{algorithm}
\usepackage{algorithmic}
\usepackage{tabularx}
\usepackage{multirow}

\begin{document}

\begin{frontmatter}

\title{Artwork Protection Against Neural Style Transfer Using Locally Adaptive Adversarial Color Attack}

\author[A]{\fnms{Zhongliang}~\snm{Guo}\orcid{0000-0002-6025-3021}\thanks{Corresponding Author. Email: zg34@st-andrews.ac.uk}}
\author[B]{\fnms{Junhao}~\snm{Dong}}
\author[A]{\fnms{Yifei}~\snm{Qian}}
\author[A]{\fnms{Kaixuan}~\snm{Wang}}
\author[A]{\fnms{Weiye}~\snm{Li}}
\author[A]{\fnms{Ziheng}~\snm{Guo}}
\author[A]{\fnms{Yuheng}~\snm{Wang}}
\author[C]{\fnms{Yanli}~\snm{Li}}
\author[A]{\fnms{Ognjen}~\snm{Arandjelovi\'c}}
\author[A]{\fnms{Lei}~\snm{Fang}}
\address[A]{University of St Andrews}
\address[B]{Nanyang Technological University}
\address[C]{University of Sydney}

\begin{abstract}
Neural style transfer (NST) generates new images by combining the style of one image with the content of another. However, unauthorized NST can exploit artwork, raising concerns about artists' rights and motivating the development of proactive protection methods.
We propose Locally Adaptive Adversarial Color Attack (LAACA), empowering artists to protect their artwork from unauthorized style transfer by processing before public release.
By delving into the intricacies of human visual perception and the role of different frequency components, our method strategically introduces frequency-adaptive perturbations in the image.
These perturbations significantly degrade the generation quality of NST while maintaining an acceptable level of visual change in the original image, ensuring that potential infringers are discouraged from using the protected artworks, because of its bad NST generation quality.
Additionally, existing metrics often overlook the importance of color fidelity in evaluating color-mattered tasks, such as the quality of NST-generated images, which is crucial in the context of artistic works.
To comprehensively assess the color-mattered tasks, we propose the Aesthetic Color Distance Metric (ACDM), designed to quantify the color difference of images pre- and post-manipulations.
Experimental results confirm that attacking NST using LAACA results in visually inferior style transfer, and the ACDM can efficiently measure color-mattered tasks.
By providing artists with a tool to safeguard their intellectual property, our work relieves the socio-technical challenges posed by the misuse of NST in the art community.
\end{abstract}

\end{frontmatter}

\section{Introduction}

Neural style transfer (NST)~\citep{gatys2015neural} is widely adopted in computer vision, where the distinctive stylistic elements of one image are algorithmically merged with the content features of another image using neural networks.
While NST opens new avenues in artistic expression and digital image processing, it poses risks of misuse, particularly in the unauthorized use of curated artworks uploaded online.
This concern has been raised by the British Broadcasting Corporation (BBC)~\citep{chris2023ai}, reporting that ``many artists and photographers say they (a company named Stability AI) use their work without permission''.
Research efforts have been put into using the neural steganography techniques for post-violation accountability in post-NST images~\citep{garg2023neural}, but, to our knowledge, we witness an absence of proactive approaches that can protect artworks from unlawful replication and manipulation induced by NST before any financial and reputational damages occur.

\begin{figure}[tb]
    \centering
    \includegraphics[width=0.89\columnwidth]{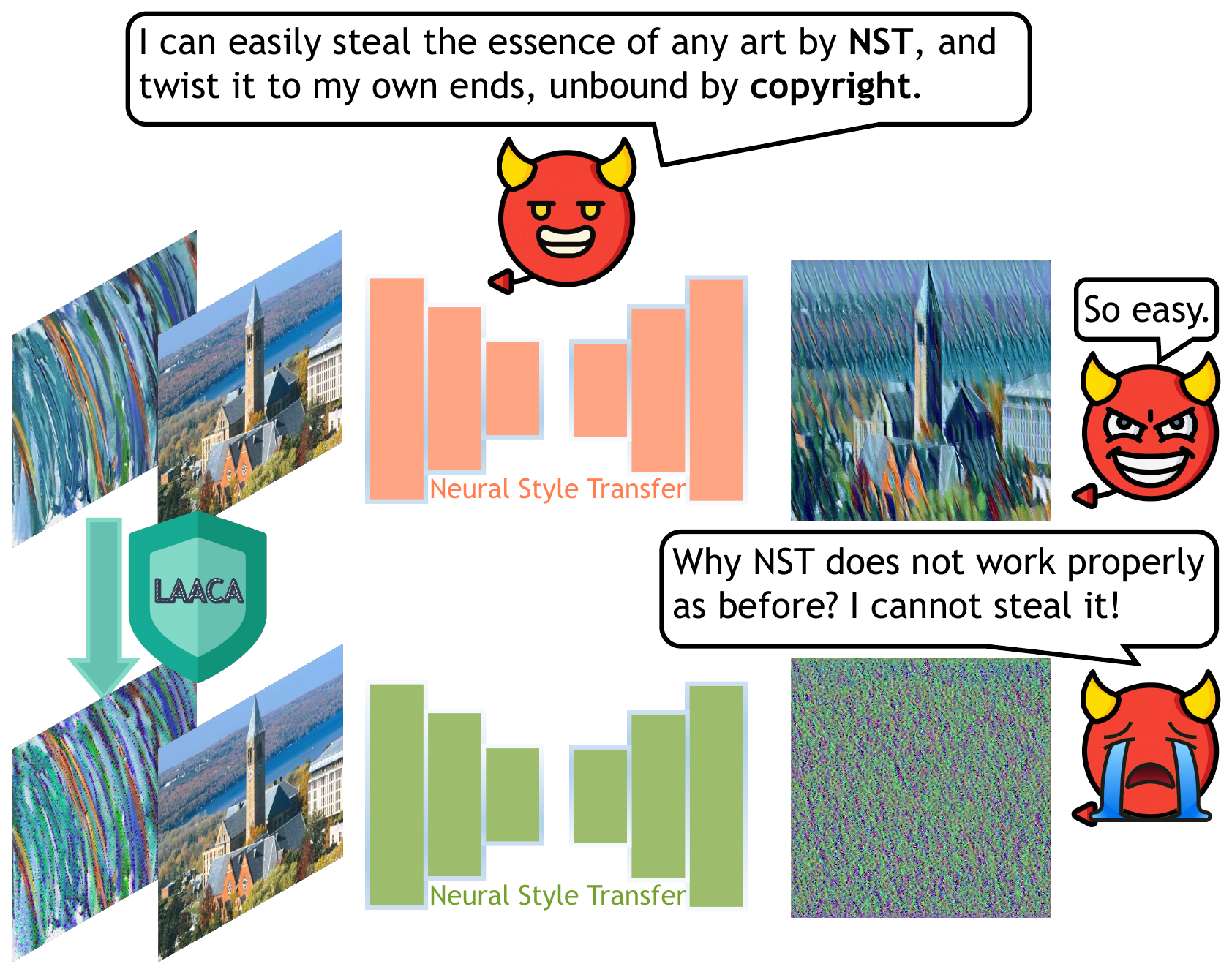}
    \caption{Role of our proposed method in preventing copyright infringement raised by unauthorized NST.}
    \label{fig:arch}
\end{figure}

\begin{figure*}[tb]
\centering
    \includegraphics[width=\textwidth]{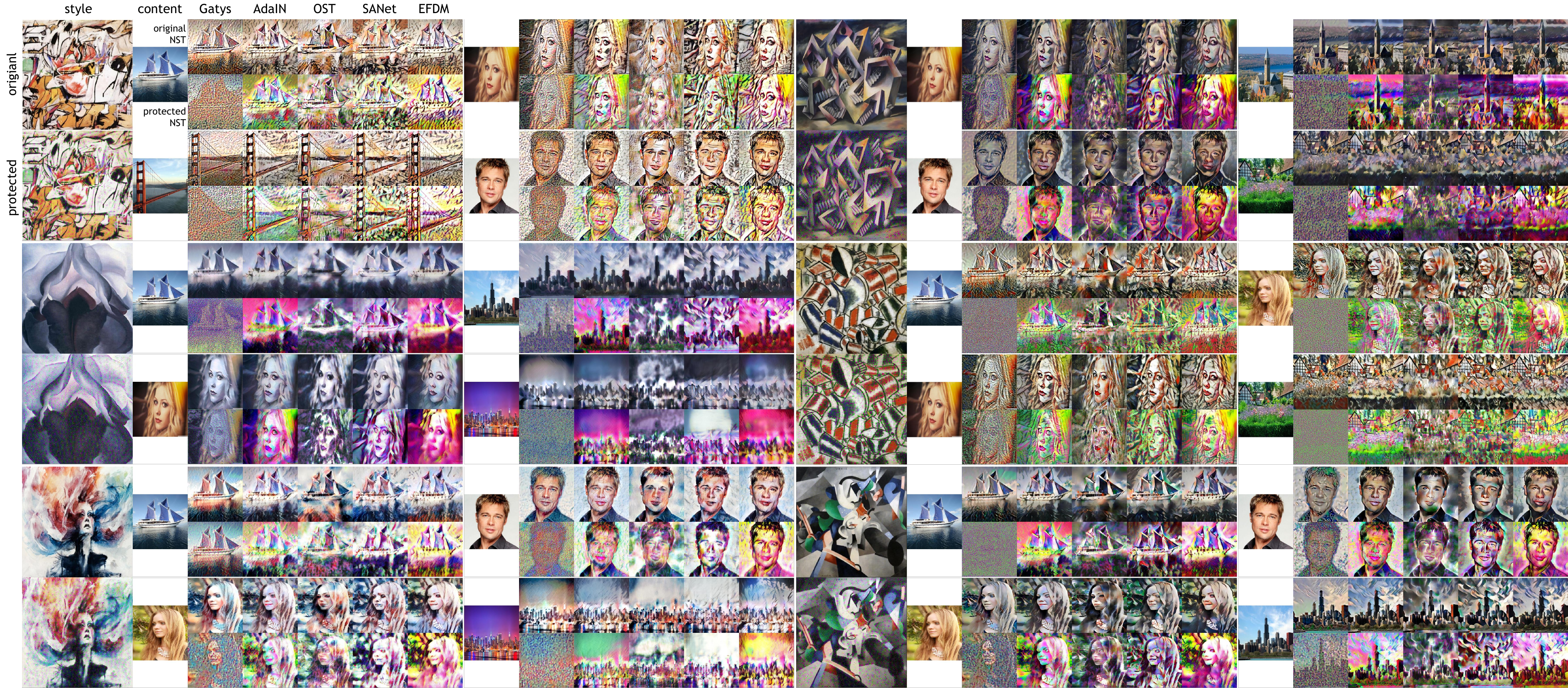}
    \caption{
    Adversarial examples against 5 NST methods on different style images and content images.
    For each item, on the far left, we exhibit the foundational images: the original style image at the top while the post-attack style image below it.
    Progressing to the right, the sequence is organized into four distinct groups for analysis.
    Each group commences with the content image, which provides the subject for the NST. Subsequent columns within each group depict the results of various NST methods (from left to right, they are Gatys~\citep{gatys2015neural}, AdaIN~\citep{huang2017arbitrary}, OST~\citep{lu2019closed}, SANet~\citep{park2019arbitrary}, and EFDM~\citep{zhang2022exact}).
    The top row across these groups showcases results from the original NST, the bottom row, in contrast, illustrates the post-attack NST outcomes.
    Specifically, most images displayed colors that were not visually present in the original style image and post-attack style image.
    Also, the textures in most images also suffer from disturbance.}
    \label{fig:introexample}
\end{figure*}

Adversarial attacks~\citep{szegedy2013intriguing}, a concept primarily explored in machine learning security, have shown promise in subtly altering input data to mislead neural networks.
Several studies have demonstrated the effectiveness of adversarial attacks in various domains~\cite{goodfellow2014explaining,liu2022segment,zhao2023prompt,zhao2024survey,zhao2024exploring}.
Inspired by the disruptive effect of adversarial attacks on machine learning-based systems, we propose to leverage this technique for artwork protection in the context of NST. By strategically embedding specific patterns or "adversarial perturbations" into digital artworks, we aim to systematically disrupt the unauthorized use of original artworks by AI models in advance. This approach offers a more robust and proactive defense mechanism compared to traditional methods like watermarking, as it directly targets the vulnerabilities of neural networks used in NST.

Color plays a crucial role in the perception and aesthetics of visual art~\citep{livingstone1988art,zeki2000inner}. In the context of NST, color consistency is a fundamental aspect of style transfer algorithms~\citep{gatys2016image,li2017universal}. However, most existing reference-based image quality assessment metrics focus on image structure~\citep{wang2004image} or semantics~\citep{heusel2017gans,prashnani2018pieapp,zhang2018unreasonable}, with limited attention given to color. This oversight leads to a lack of evaluation metrics specifically designed for color-sensitive tasks like NST. To address this issue, we propose the Aesthetic Color Distance Metric (ACDM), a novel metric that quantifies the color changes of images after undergoing certain transformations. By capturing color-related properties, 
ACDM provides a more comprehensive evaluation of color changes between pre- and post-manipulated images, which will also be helpful to exhaustively evaluate the proposed artwork protection method.

In light of the practical restrictions of artwork protection, we identify three main desiderata for image-level alterations tailored for NST:
(a) acceptable perceptibility to the human eye that ensures the artwork's visual integrity,
(b) effectiveness in disrupting the generation quality of most NST methods, and
(c) a generic solution applicable to broad-spectrum of artworks.
To address these requirements, we propose the Locally Adaptive Adversarial Color Attack (LAACA), a method that integrates adversarial techniques directly into the digital artwork creation process.
LAACA leverages a frequency domain filter to divide the image into high-frequency and low-frequency content zones, and clips the perturbations in the high-frequency zone during each iteration of the attack. This approach ensures the visual integrity of the attacked images while effectively disrupting the color features and local texture details of the post-attacked NST images.
Figure~\ref{fig:introexample} demonstrates the impact of LAACA on NST outputs.

The main contributions of this work are as follows:
\begin{itemize}
\item We propose LAACA, a novel artwork protection method that proactively safeguards digital image copyrights by disrupting the NST generation through the addition of visually imperceptible perturbations to the input style image prior to the NST process.
\item To address the limitations of existing image quality assessment metrics in evaluating color-mattered tasks, we introduce ACDM, a new metric that quantifies the color changes of images after undergoing certain transformations.
\end{itemize}

\section{Related Works}
In this section, we review the relevant literature in the fields of neural style transfer and adversarial attacks. We first discuss the evolution of NST algorithms, from the seminal work of Gatys et al.~\citep{gatys2015neural} to more recent advancements in Arbitrary Style Transfer. We then delve into the development of adversarial attacks, highlighting the shift towards a frequency domain perspective and the progress made in applying adversarial attacks to domains beyond image classification. Finally, we identify the research gap in adversarial attacks specifically targeting NST and position our work in the context of this gap.

\paragraph{Neural style transfer.}
Neural Style Transfer (NST) witnessed a foundational advancement with the work of~\citet{gatys2015neural}, which enabled the transfer of artistic style characteristics from one image to another through an iterative optimization process using the Gram Matrix.
Building on this seminal work, subsequent research in NST explored alternatives to the Gram Matrix, offering improved stylization outcomes~\citep{gatys2016preserving,kalischek2021light,luan2017deep,risser2017stable}.
A significant evolution in NST was the transition to non-iterable forms, known as Arbitrary Style Transfer (AST).
A key development in this area was Adaptive Instance Normalization (AdaIN)~\citep{huang2017arbitrary}, which simplified the style transfer process by training a decoder with fused statistical features of the style and content images.
Furthermore,~\citet{lu2019closed} offered a closed-form solution for NST, further streamlining the style transfer process.
\citet{park2019arbitrary} integrated the attention mechanism into NST, enhancing the effectiveness of style transfer.
Notably,~\citet{zhang2022exact} updated the matching function in AdaIN by introducing Exact Feature Distribution Matching (EFDM), allowing for much better AST.
It is important to highlight that our work does not aim to alter the parameters of NST algorithms; instead, we focus on manipulating the input style images to disrupt the style transfer process, offering a novel perspective on adversarial attacks in the context of NST.

\paragraph{Adversarial attack.}
The exploration of adversarial attacks against neural networks was pioneered by~\citet{szegedy2013intriguing}, who underscored the susceptibility of classification neural networks to perturbations in the input.
Following this groundbreaking work,~\citet{goodfellow2014explaining} introduced one-shot adversarial perturbations by leveraging the gradients of neural networks to deceive classification models.
\citet{carlini2017towards} proposed the first successful targeted attack on classification models trained with ImageNet~\citep{deng2009imagenet}.
\citet{madry2017towards} iteratively constrained image perturbations, allowing for more efficient convergence.
\citet{moosavi2017universal} proposed Universal Adversarial Perturbation, which can fool models with a single perturbation for arbitrary data.
By introducing momentum in iterations,~\citet{dong2018boosting} further increased the transferability of adversarial samples.

A notable shift in the approach to adversarial attacks has been towards a frequency domain perspective, focusing on the role of frequency composition in the effectiveness and perceptibility of adversarial perturbations.
\citet{guo2018low} highlighted that solely using low-frequency noise can reduce computational costs for black-box attacks.
Furthermore,~\citet{maiya2021frequency} offered that the frequency of noise in adversarial attacks is not strictly high or low but is related to the dataset.
Advancing this inquiry,~\citet{jia2022exploring} explored generating perturbations in the frequency domain.
\citet{wang2023lfaa} employed a conditional decoder to generate low-frequency perturbations, enabling a fast targeted attack.
These developments suggest that considering adversarial attacks from a frequency domain standpoint could provide a more refined understanding and potentially enhance the effectiveness of attacks.
Building upon these seminal advancements in adversarial attacks, the field has progressed into other domains of artificial intelligence~\citep{dong2021visually,dong2023restricted,guo2023white,zhao2024defending,zhao2024universal}.

To the best of our knowledge, there is only one method that attacks NST by disrupting \emph{content images}, with no direct exploration of altering style images.
The mentioned content-disruptive method, Feature Disruptive Attack (FDA)~\citep{ganeshan2019fda}, manipulates the intermediate features of \emph{content images} mapped by a neural network, resulting in distorted content in post-NST images while the applied style remains unchanged.
However, the visual difference between pre- and post-attack images by FDA is slightly obvious.
In contrast, our work focuses on adding imperceptible perturbations to \emph{style images}, which results in significantly degraded post-NST images, regardless of the content images used.
This content-independent method opens up new possibilities for adversarial attacks against unauthorized-NST usage, offers a more flexible and generalizable approach to disrupting NSTs.

\section{Methodology}
In this section, we first define the problem of artwork protection against NST in the adversarial attack framework via a simple yet unified formulation and propose our method.
Additionally, we design a color-based metric named Aesthetic Color Distance Metric (ACDM) to assess the artistic style difference, which complements the existing Image Quality Assessment (IQA).

\subsection{Problem Definition}
We commence with a style image $\boldsymbol{x}\in \mathbb{R}^{C \times H\times W}$ from a set of style images $\mathcal{X}$, where $C$, $H$, and $W$ denote the channels, height, and width of the image, respectively. Similarly, a content image $\boldsymbol{y}\in \mathbb{R}^{C\times H\times W}$ is selected from a content image set $\mathcal{Y}$.

The function $\boldsymbol{g}=\textnormal{NST}(\boldsymbol{x},\boldsymbol{y})$ represents the neural style transfer process, which amalgamates the style of image $\boldsymbol{x}$ with the content of image $\boldsymbol{y}$. The output $\boldsymbol{g}$ denotes the resultant style-transferred image.

We introduce $\boldsymbol{x}^\ast$ as the protected style image generated from $\boldsymbol{x}$,
where $\boldsymbol{x}^\ast = \boldsymbol{x} + \boldsymbol{\delta}$, and the difference vector $\boldsymbol{\delta} = \boldsymbol{x}^\ast - \boldsymbol{x}$ is the perturbation designed to disrupt NST. The essence of disrupting NST lies in creating a protected style image $\boldsymbol{x^*}$, visually similar to $\boldsymbol{x}$, yet significantly altering the NST generation combined with an arbitrary content image $\boldsymbol{y}$. 
To make changes visually imperceptible, the perturbation is restricted in an $\ell_p$ norm, denoted as $\|\boldsymbol{\delta}\|_p\leq\epsilon$, where $\epsilon$ is the defined budget of perturbations. Thus, the problem is:
\begin{equation}\label{eq:optim:dual}
\begin{aligned}
    &\arg\max_{\boldsymbol{\delta}}
\, \mathbb{E}_{\boldsymbol{y}\sim P_{\mathcal{Y}}}[D(\boldsymbol{g},\boldsymbol{g^*})]\quad s.t.\, \|\boldsymbol{\delta}\|_p \leq \epsilon\,,
\end{aligned}
\end{equation}
 we assume $D$ is to measure the human perceptual distance;
the post-protection NST output is $\boldsymbol{g}^\ast = \textnormal{NST}(\boldsymbol{x}^\ast,\boldsymbol{y})$;
the expectation is taken with the content images' population distribution $P_{\mathcal{Y}}$.

Previous unauthorized NST prevention methods, like neural steganography or watermarking, offer post-violation accountability but rely on detecting infringement after the fact. Conversely, the adversarial attack method proactively introduces imperceptible perturbations that degrade NST output quality, deterring potential infringers by rendering the resulting images unsuitable for their intended purpose. the supplementary material illustrates
the difference of those different technique approaches.

\subsection{Locally Adaptive Adversarial Color Attack}

\begin{figure}[tb]
\centering
    \includegraphics[width=\columnwidth]{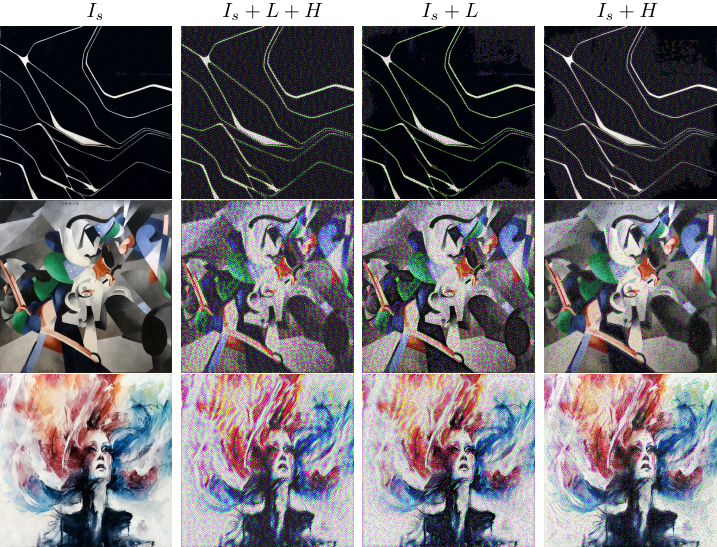}
    \caption{
    This figure illustrates various perturbation patterns applied to various style images, among perturbed images, ``$I_s + H$'' remains the best visual integrity. To be specific, $I_s$ is the clean style image; $L$ means low-frequency components of perturbations; $H$ means high-frequency components of perturbations. Those perturbations are generated by our method with $k=4$, $\alpha=8$, $\epsilon=80$, and $T=100$. We separate the different frequency components by the kernel in Equation~\ref{eq:gaussian} with $k=4$.}
    \label{fig:pertubablation}
\end{figure}

Building upon the Equation~\ref{eq:optim:dual}, there will be a challenge:
considering the extensive variety of content images $\boldsymbol{y} \in P_{\mathcal{Y}}$ for NST, it becomes impractical to enumerate and process every content and style image combinations for each single style image. For example, using the ImageNet~\citep{deng2009imagenet} as a benchmark, we recognize that a representative content image subset would include around 1,000 categories, and each category has a significant in-class variance. We assume that sampling at least 10 images from each class would slightly cover this in-class variance, leading to a minimum of 10,000 samples for the subset. This approach poses a substantial computational challenge, particularly for iterative NST methods where time costs are significant.

To address this challenge, we try to make a paradigm shift to a method that is only reliant on style image $\boldsymbol{x}$. In this new framework, we introduce an amortized encoder, denoted as $f$, coupled with a surrogate loss function, $J$.
The encoder $f$ is designed to take an input style image $\boldsymbol{x}$ and output its content-less style representation.
This representation is a distilled essence of the style image's characteristics, capturing features that define its unique artistic style.
The surrogate loss function $J$ then measures the disparity between two style representations generated by $f$. This approach allows us to approximate the
$\mathbb{E}_{\boldsymbol{y}\sim P_{\mathcal{Y}}}[D(\operatorname{NST}(\boldsymbol{x},\boldsymbol{y}), \operatorname{NST}(\boldsymbol{x}^\ast,\boldsymbol{y}))]$
with $J(f(\boldsymbol{x}),f(\boldsymbol{x}^\ast))$, which is content-independent. Therefore, our task is to maximize the difference between the style representations of the original and protected style images, while ensuring the perturbation $\boldsymbol{\delta}$ is within a pre-defined budget limit $\epsilon$, resulting acceptable visual integrity:
\begin{equation}\label{eq:optim:final}
    \begin{aligned}
        \arg\max_{\boldsymbol{\delta}}J(f(\boldsymbol{x}),f(\boldsymbol{x}^\ast)),\,s.t.\, \|\boldsymbol{\delta}\|_p \leq \epsilon\,.
    \end{aligned}
\end{equation}

Inspired by previous works~\citep{legge1980contrast,wang2023lfaa}, we restrict the perturbation $\boldsymbol{\delta}$ to be frequency-adaptive such that the visual effect is better preserved. This is motivated by the observation that 
the high correlation between different frequency components and visual effects. 
This finding guides us to embed adversarial perturbations within the high-frequency areas of the style image. This idea is demonstrated in Figure~\ref{fig:pertubablation}, where restricting perturbations to the high-frequency zone maintains higher visual integrity compared to other patterns. More formally, denote the pixel set of high-frequency components of an image $\boldsymbol{x}$ as $M({\boldsymbol{x}}) \subseteq \{(i, j)\mid i = 1,\ldots, H, j = 1,\ldots, W\}$, and its complement ${\neg M(\boldsymbol{x})}$ becomes the corresponding low-frequency pixel set.
Thus, the problem is formulated as:
\begin{equation}\label{eq:optim:single}
    \begin{aligned}
        &\arg\max_{\boldsymbol{\delta}}J(f(\boldsymbol{x}),f(\boldsymbol{x}^\ast)),\\
         &s.t.\, \|\boldsymbol{\delta}\|_p \leq \epsilon \; \text{and } \boldsymbol{\delta}[i, j] =0, \, \text{for } (i,j) \in \neg M(\boldsymbol{x}).
    \end{aligned}
\end{equation}

\paragraph{Frequency separator.}\label{sec:freqfilter}
We employ a low-pass Gaussian filter to separate different frequency components from an image:
\begin{equation}\label{eq:gaussian}
    G_k\left(i,j\right)=\frac{1}{2\pi k^2}e^{-\frac{i^2+j^2}{2k^2}},
\end{equation}
where $G_k(i,j)$ denotes the value of the Gaussian kernel at position $(i,j)$. The standard deviation of the kernel, $\sigma$, is determined by $k$. The kernel size is $(4k+1)\times (4k+1)$. The output vector from this kernel is the low-frequency components of the image. By using the above frequency separator, the high-frequency components of the style image $\boldsymbol{x}$ become:
\begin{equation}\label{eq:position-set}
\begin{aligned}
    M(\boldsymbol{x}) = \{ (h,w)&|\boldsymbol{x}-G_{k}(h,w)>0;\\
    &h,w\in \mathbb{N}; 1<h\leq H, 0<w\leq W\},
\end{aligned}
\end{equation}
where $\mathbb{N}$ represents the set of natural number. The pixels in an image $\boldsymbol{x}$ belonging to $M(\boldsymbol{x})$ are denoted as high-frequency zone, and pixels belonging to $\neg M(\boldsymbol{x})$ is denoted as low-frequency zone.

\begin{algorithm}[tb]
    \caption{Locally Adaptive Adversarial Color Attack (LAACA)}
    \label{alg:laaca}
    \textbf{Input}: A style transfer encoder $f$ with style loss function $J$; a real style image $\boldsymbol{x}$; a Gaussian kernel $G_k$ with kernel size $k$\\
    \textbf{Parameter}: The attack step size $\alpha$ ; $\ell_{\infty}$-norm perturbation radius $\epsilon$; iterations $T$\\
    \textbf{Output}: Attacked style image $\boldsymbol{x}^\ast$\vfill
    \begin{algorithmic}[1]
        \STATE $clamp_m^n$ restricts a value to be within the range $\left[m, n\right]$
        \STATE 
        % To avoid loss is 0 at the beginning, 
        randomly generate $\boldsymbol{\delta_0}$ in $\left[0,2\right]$ to avoid the gradient is 0 in loops
        \STATE $\boldsymbol{x}^\ast_0=clamp_0^{255}\left[\boldsymbol{x}+M(\boldsymbol{\delta_0})\right]$
        \FOR{$t=0$ to $T-1$}
        \STATE $\boldsymbol{x}_t^\ast$ requires gradient
        \STATE Input $\boldsymbol{x}_t^\ast$ and $\boldsymbol{x}$ to $f$ and obtain the gradient\\$\nabla_{\boldsymbol{x}} J\left(f\left(\boldsymbol{x}^\ast_t\right),f\left(\boldsymbol{x_{\null}^{\null}}\right)\right)$
        \STATE Update $\boldsymbol{x}_t^\ast$ by accumulating the signed gradient\\
        $\boldsymbol{x}_t^\ast=\boldsymbol{x}_t^\ast+\alpha\cdot sign\left[\nabla_{\boldsymbol{x}} J\left(f\left(\boldsymbol{x}^\ast_t\right),f\left(\boldsymbol{x_{\null}^{\null}}\right)\right)\right]$
        \STATE Get the perturbation and apply the mask on it\\
        $clamp_{-\epsilon}^{\epsilon}\boldsymbol{\delta_t}\left[\neg M(\boldsymbol{x})\right]=0$
        \STATE Update $\boldsymbol{x}_{t+1}^\ast$ by the masked perturbation\\
        $\boldsymbol{x}_{t+1}^\ast=clamp_0^{255}\left[ \boldsymbol{x}+\boldsymbol{\delta_t}\right]$
        \ENDFOR
        \STATE \textbf{return} $\boldsymbol{x}_t^\ast$
    \end{algorithmic}
\end{algorithm}

\paragraph{Encoder.}
For the amortized encoder $f$, we utilize a pre-trained VGG~\citep{simonyan2014very}, using its several layer outputs as feature extraction encoders. This choice is inspired by established arbitrary style transfer methods that effectively extract images' style representation from intermediate network layers. Each layer of the network is denoted as $l$, collectively denoted as a set $L$, $f^{l}(\boldsymbol{x})$ indicates the mapped result of intermediate layer $l$ of the style image $\boldsymbol{x}$.

\paragraph{Color disruptive loss function.}
As for the surrogate loss function $J$, our goal is to measure aspects of the neural network's intermediate layer mappings that represent color.
This leads us to consider the mean $\mu$ and standard deviation $\sigma$, which are important in the neural network feature representation in terms of NST.
Evidenced by \citet{zhang2022exact}, who tested the influence of $\mu$ and $\sigma$, when only matching the $\mu$ of content representations with $\mu$ of style representations, the color of the post-NST image will be the same as that of its style counterpart; in contrast, when only matching $\sigma$, the texture will be similar. That is, $\mu$ represents the color, while the $\sigma$ represents the contrast and texture variations, both of which significantly contribute to an image's style. Therefore, we design the surrogate loss function targeting those two statistics:
\begin{equation}\label{eq:loss}
\begin{aligned}
    J\big(f(\boldsymbol{x}),f(\boldsymbol{x}^\ast)\big) =
     \sum_{l\in L} \big(\big(&\mu(f^{l}(\boldsymbol{x}))-\mu(f^{l}(\boldsymbol{x}^\ast))\big)^2\\+\big(&\sigma(f^{l}(\boldsymbol{x}))-\sigma(f^{l}(\boldsymbol{x}^\ast))\big)^2\big)\,,
\end{aligned}
\end{equation}
where $\mu$ is a function to get the mean of feature in each channel, and $\sigma$ is a function to get the standard deviation of feature in each channel.
By focusing on these aspects, function $J$ can effectively guide our method in disrupting the generation of the neural style transfer.

\paragraph{Generation of perturbation.}
Algorithm~\ref{alg:laaca} outlines the protection method transforming $\boldsymbol{x}$ to $\boldsymbol{x}^\ast$.\footnote{Our code is available at \url{https://github.com/ZhongliangGuo/LAACA}.} We employ an iterative approach with an $\ell_\infty$ norm constraint, denoted as $\|\boldsymbol{x} - \boldsymbol{x}^\ast\|_\infty \leq \epsilon$. before the for loop, we randomly generate a small noise $\boldsymbol{\delta_0}$ to avoid the gradient is 0 in loops. For each iteration, the gradient of the loss function with respect to the input is computed, and the perturbations are updated in the direction maximizing the loss value, with a step size of $\alpha$. The perturbations are then clipped to maintain the $\ell_\infty$ constraint.

\subsection{Aesthetic Color Distance Metric}
At present, there is no image distance metric specifically designed for distinguishing image differences affected by style from color perspective, i.e., the color consistency is often overlooked in existing research; the commonly used generic metrics are not well-suited for this task either, as evidenced by the experimental results summarized in Table~\ref{tab:eval:acdm}.
To address this limitation,
we propose a new metric, Aesthetic Color Distance Metric (ACDM),\footnote{Our code is available at \url{https://github.com/ZhongliangGuo/ACDM}.} in the LAB color space, 
which is constructed to be perceptually uniform and more closely aligns with human visual perception compared to the RGB space~\citep{ruderman1998statistics}. The LAB color space consists of three channels: L represents lightness, ranging from 0 to 100, A and B represent color opponents, with A ranging from -128 (\textcolor{green}{$\bullet$}) to 127 (\textcolor{red}{$\bullet$}) and B ranging from -128 (\textcolor{blue}{$\bullet$}) to 127 (\textcolor{yellow}{$\bullet$}).

Given two images $\boldsymbol{z}_1, \boldsymbol{z}_2 \in \mathbb{R}^{C \times H \times W}$ in the LAB color space, where $C=3$ denotes the number of color channels (L, A, B), and $H, W$ denote the image height and width, respectively. For each color channel $c \in \{L, A, B\}$, we compute the histogram of values. 
The histogram represents the distribution of pixel values for each color channel, and is considered an effective method to characterize the color composition of an image.

The number of bins $N_c$ for each channel is determined by taking the square root of the difference between the maximum and minimum values of that channel:
\begin{equation}
    N_c =\left \lceil \sqrt{{\max}_c-{\min}_c} \right \rceil
\end{equation}
where $\min_c$ and $\max_c$ denote the possible value range of channel $c$. Denote the bin width as $h = \tfrac{\max_c-\min_c}{N_c}$, the corresponding bins are $\{B_i\}_{i=1}^{N_c}$, where $B_i= \left [{\min}_c + (i-1)h, {\min}_c + ih\right)$ for $i =1,\ldots, N_c-1$, and $B_{N_c} = \left [{\min}_c + (N_c-1)h, {\max}_c\right)$.

Let $\mathcal{H}_c: \mathbb{R}^{H \times W} \to \mathbb{R}^{N_c}$ denote the function that maps a channel image to its corresponding $N_c$-dimensional frequency count vector. To enable a meaningful comparison between the color distributions of the two images, we normalize the histogram vectors of both images $\boldsymbol{z}_1$ and $\boldsymbol{z}_2$ by dividing each histogram by its own sum:
\begin{equation}
    \begin{aligned}
        \hat{\boldsymbol{h}}_c(\boldsymbol{z}_i) = \frac{\mathcal{H}_c(\boldsymbol{z}_{i,c})}{\sum_{j=1}^{N_c} \mathcal{H}_c(\boldsymbol{z}_{i,c})_j}\,.
    \end{aligned}
\end{equation}
This normalization step converts the histogram vectors into probability distributions, ensuring that the resulting normalized histograms $\hat{\boldsymbol{h}}_c(\boldsymbol{z}_1)$ and $\hat{\boldsymbol{h}}_c(\boldsymbol{z}_2)$ have a total sum equal to 1. By representing the color distributions as probability distributions, we can effectively capture the relative frequencies of pixel intensities in each channel, facilitating a fair comparison between the two images.

\paragraph{Earth Mover's Distance based metric.} To measure the color difference between $\boldsymbol{z}_1$ and $\boldsymbol{z}_2$, we employ the Earth Mover's Distance (EMD) between their normalized histograms for each channel. The EMD calculates the minimum cost required to transform one probability distribution into another. By using the EMD, we consider not only the absolute differences between corresponding histogram bins but also the overall shape and structure of the distributions. 
This is particularly important in the context of color distributions, as the EMD can capture perceptually meaningful differences that simple bin-wise comparisons may overlook. Moreover, the EMD takes into account the ground distance between bins, which allows for a more nuanced comparison of the color distributions.

When the two distributions are 1-D vectors, the distance can be solved with a closed-form solution~\citep{kolouri2018sliced,lin2024semi}:
\begin{equation}
\begin{aligned}
    &\mathcal{D}_c(\boldsymbol{z}_1,\boldsymbol{z}_2) = \frac
    {\sum_{n=1}^{N_c}|\mathcal{F}(\hat{\boldsymbol{h}}_c(\boldsymbol{z}_1))_n-\mathcal{F}(\hat{\boldsymbol{h}}_c(\boldsymbol{z}_2))_n|}{N_c-1}\,.
\end{aligned}
\end{equation}
where $\mathcal{F}(\hat{\boldsymbol{h}}_c(\boldsymbol{z}_i))_n = \sum_{j =1\ldots n}\hat{\boldsymbol{h}}_c(\boldsymbol{z}_i)_j$ is the cumulative probability mass up to the $n$-th bin.
To scale the data for better interpretability, we adopt a max-min normalization with a theoretical maximum difference of two distributions.
By employing the EMD, we obtain a robust and semantically meaningful measure of the color difference between the two images in each perceptual color channel.
Finally, we obtain the overall color difference score by summing the differences across all three channels:

\begin{equation}
    \textnormal{ACDM}(\boldsymbol{z}_1, \boldsymbol{z}_2) \quad=\ \sum_{c\in \{L,A,B\}}\mathcal{D}_c(\boldsymbol{z}_1, \boldsymbol{z}_2).
\end{equation}

\paragraph{Discussion on distance metrics.}
Consider a toy example of histograms from one channel of 3 images, each with 4 bins. Suppose the normalized histograms of these images are:
$A = [0.0, 1.0, 0.0, 0.0]$,
$B = [0.2, 0.3, 0.5, 0.0]$,
$C = [0.2, 0.3, 0.0, 0.5]$.
Histograms $B$ and $C$ have the same bin values but in a different order. Intuitively, the distribution of $B$ ought to be more similar to $A$ because the position of its 0.5 entry is closer to that of 1.0 in $A$ compared with $C$.
However, when we use $\ell_1$ loss ($\mathcal{H}_1$), $\ell_2$ loss ($\mathcal{H}_2$), Cross Entropy ($\mathcal{H}_3$), Cosine Similarity ($\mathcal{H}_4$) or Euclidean Distance ($\mathcal{H}_5$) to compare these vectors, we find that $B$ and $C$ have the same loss values when compared to $A$:
$\operatorname{\mathcal{H}_1}(A,B) = \operatorname{\mathcal{H}_1}(A,C)=0.35$,
$\operatorname{\mathcal{H}_2}(A,B) = \operatorname{\mathcal{H}_2}(A,C)=0.195$,
$\operatorname{\mathcal{H}_3}(A,B)=\operatorname{\mathcal{H}_3}(A,C)=1.4437$,
$\operatorname{\mathcal{H}_4}(A,B)=\operatorname{\mathcal{H}_4}(A,C)=0.4867$,
$\operatorname{\mathcal{H}_5}(A,B)=\operatorname{\mathcal{H}_5}(A,C)=0.8832$,
which indicates that although $B$ and $C$ have different similarities to $A$, these three loss functions cannot distinguish between them.
On the other hand, if we use EMD to compare these vectors, we find that:
$\operatorname{EMD}(A,B)=0.7,\operatorname{EMD}(A,C)=1.2.$

EMD provides a more intuitive and accurate measure of the difference between the distributions.
The smaller EMD value between $A$ and $B$ reflects that their high-value bins are concentrated in the same region, and only a slight movement of some pixels is needed to match them perfectly. In contrast, the larger EMD value between $A$ and $C$ indicates that more pixels need to be redistributed among the bins to match distributions.

\section{Results}
\subsection{Experimental Setup}
We use the normalized VGG-19~\citep{gatys2016image} as our encoder, consistent with NST methods like AdaIN and EFDM, capturing intermediate layer outputs from ${relu1\_1, relu2\_1, relu3\_1, relu4\_1}$. We set $k=4$, $\alpha=8$, $\epsilon=80$, $T=100$ to balance visual integrity and attack effectiveness, with hyperparameter discussions in the ablation studies.
Content images are sourced from MS-COCO~\citep{lin2014microsoft}, and style images are from WikiArt~\citep{painter-by-numbers}.

We target five popular NST methods:
Gatys~\citep{gatys2015neural},
\footnote{\url{https://pytorch.org/tutorials/advanced/neural_style_tutorial.html}}
AdaIN~\citep{huang2017arbitrary},
\footnote{\url{https://github.com/naoto0804/pytorch-AdaIN}}
OST~\citep{lu2019closed},
\footnote{\url{https://github.com/boomb0om/PyTorch-OptimalStyleTransfer}}
SANet~\citep{park2019arbitrary},
\footnote{\url{https://github.com/GlebSBrykin/SANET}}
and
EFDM~\citep{zhang2022exact},
\footnote{\url{https://github.com/YBZh/EFDM}}
representing various approaches in the NST domain.
For all NST methods, we set the image size as $512\times512$, for Gatys, we follow the initial setting, setting $style\_weight=1e^6, content\_weight=1,epochs=500$; For OST, $\alpha=0.6$; For AdaIN, SANet and EFDM, we apply the default parameter $\alpha=1$ which was discussed in their paper.
We randomly sample around 300 pairs of style and content images to evaluate original and protected artworks, pre-and post-protection NST images.

As no existing attack methods are specifically designed for style images of NST, we consider the Universal Adversarial Perturbation (UAP)~\citep{moosavi2017universal} as a baseline due to its wide applicability in adversarial settings. To align with our method, the $\ell_\epsilon$ norm of two baselines is also set as 80.

To further evaluate the performance of our method in real-world scenarios, where images are often compressed or downscaled for efficient distribution, we simulate common image degradation techniques by applying JPEG compression (retain 75\% quality) and Gaussian blur (kernel in Equation~\ref{eq:gaussian} with $k=3$) to the test images.

\subsection{Evaluation for ACDM} 
Color is a key indicator for image style differences, often overlooked by existing metrics. Our proposed Color-based metric, ACDM, effectively distinguishes between style differences from color perspective.

Due to the current lack of well-annotated datasets for color-centric tasks, we validate ACDM in the Neural Style Transfer (NST) context.
We hypothesize that positive pairs (same style, different content) should have higher color correlations than negative pairs (same content, different styles). We expect smaller ACDM scores for positive pairs and larger scores for negative pairs.

We sample 10,000 pairs from MS-COCO (content) and WikiArt (style), perform style transfer using EFDM, and compare ACDM with two popular metrics Structural Similarity Index Measure for color image (SSIMc)~\citep{wang2004image} and Learned Perceptual Image Patch Similarity (LPIPS)~\citep{zhang2018unreasonable}, where SSIMc is a SSIM's variant considering the color information. It is worth noting that we use the LPIPS with VGG, aligning with the convention in NST domain. The evaluation results are shown in Table~\ref{tab:eval:acdm}.

\begin{table}[hb]
\centering
\caption{Evaluation on effectiveness of ACDM compared with SSIMc and LPIPS, $\uparrow$/$\downarrow$ indicates that bigger/smaller value means better image quality.}
\label{tab:eval:acdm}
\resizebox{\columnwidth}{!}{
\begin{tabular}{ccc|c}
\toprule
IQA & positive pairs ($\mathbb{P}$) & negative pairs ($\mathbb{N}$) & change ratio $\frac{|\mathbb{P}-\mathbb{N}|}{\mathbb{P}}$ \\ \midrule
SSIMc $\uparrow$                        & 0.2871         & 0.4072         & 41.83\%  \\
LPIPS$_{\textnormal{VGG}}$ $\downarrow$  & 0.5851         & 0.5459         & 6.70\%   \\ 
ACDM $\downarrow$                       & 0.0464         & 0.2982         & \textbf{542.67\%} \\ \bottomrule
\end{tabular}
}
\end{table}

In comparison, although SSIMc considers the color information, it demonstrates a more notable ability to capture image structure, which is more related to the content of the image rather than its color style. This is evidenced by the higher SSIMc score for negative pairs, where the content is consistent, compared to positive pairs, where the content varies. This suggests that SSIMc is more sensitive to changes in image content rather than to changes in color or style. This perceptual ability of SSIMc can be leveraged as an evaluation metric for aspects other than color in our proposed Locally Adaptive Adversarial Color Attack (LAACA) method.

Furthermore, in the context of NST, LPIPS exhibits similar performance for both positive and negative pairs. This suggests that LPIPS may not be particularly sensitive to changes in either content or style when the other component remains consistent. In other words, LPIPS seems to be influenced by both content and style simultaneously, making it less discriminative when one of these factors is fixed. This observation highlights the need for a more targeted evaluation metric, such as our proposed ACDM, which can effectively capture color-related changes even when content or style is held constant.

These findings underscore the effectiveness of our proposed ACDM metric in evaluating color-mattered tasks. The substantial difference in scores between positive and negative pairs demonstrates its strong perceptual ability to capture color differences, setting it apart from other commonly used metrics.

\subsection{Results for LAACA}
We employ three evaluation methods to assess and analyze the experimental results: our proposed ACDM, SSIMc, LPIPS.
ACDM quantifies the color variations between the original and attacked images. SSIMc is used to measure the changes in image structure before and after the attack, with higher SSIMc values indicating greater structural similarity and a maximum value of 1 indicating identical images. LPIPS measure the perceptual similarity of the pre- and post-attack images. For both SSIMc and ACDM, we use a Gaussian kernel size of 11, following the default setting of SSIMc, to ensure a consistent and comparable evaluation. In the following tables, $+$/$-$ indicates that bigger/smaller value means better results.

\paragraph{SSIMc.}
Table~\ref{tab:result:ssim} presents the SSIMc scores comparing the structural similarity between the original and attacked images. Our method achieves an SSIMc score nearly double that of UAP, indicating excellent preservation of structural information. For the style-transferred images, LAACA obtains an average SSIMc score of 0.3356, better than the UAP, demonstrating its effectiveness in disrupting the style transfer process.
Under defense, LAACA's performance slightly declines. However, these results remain within an acceptable range, showcasing LAACA's robustness in real-world scenarios where images may undergo compression or blurring during distribution.
\begin{table}[hb]
\caption{SSIMc evaluation results, a higher score means better quality.} 
\label{tab:result:ssim}
\resizebox{\columnwidth}{!}{
\begin{tabular}{cccccccc}
\toprule
(\textbf{the best},          & \multirow{2}{*}{style images $^+$} & \multicolumn{6}{c}{Neural Style Transfer Methods}    \\ 
\underline{the second best}) &                               & Gatys  & AdaIN  & OST    & SANet  & EFDM   & Average $^-$\\ \midrule 
LAACA                        & \underline{0.6130}            & 0.2392 & 0.3891 & 0.3671 & 0.3150 & 0.3674 & \underline{0.3356}  \\ 
UAP        & 0.3556                        & 0.3059 & 0.4555 & 0.3121 & 0.2599 & 0.4222 & 0.3511  \\ 
JPEG 75\% Comp.              & \textbf{0.6218}               & 0.2639 & 0.3986 & 0.3757 & 0.3229 & 0.3765 & 0.3475  \\
Gaussian blur                & \textbf{0.6214}               & 0.5499 & 0.5354 & 0.5403 & 0.4704 & 0.5080 & 0.5208  \\ \bottomrule
\end{tabular}}
\end{table}

\begin{figure*}[t]
    \centering
        \includegraphics[width=0.99\textwidth]{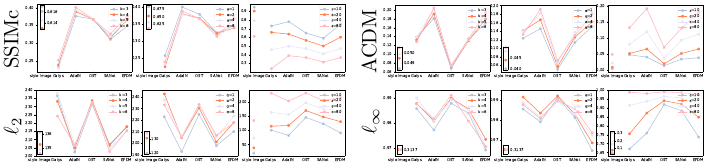}
        \caption{Ablation study results, for those labels on X-axis, they represent comparisons of pre- and post-attack images of X-axis value on a certain metric.}
        \label{fig:ablation}
\end{figure*}

\paragraph{ACDM.}
Table~\ref{tab:result:acdm} presents the ACDM scores, which measure the color difference between the original and attacked images. For the style images, our proposed LAACA method achieves an ACDM score of 0.0495, significantly better than the UAP (0.1350), indicating excellent preservation of visual color information in protected images. For the style-transferred images, LAACA obtains the highest average ACDM score, demonstrating its ability to disrupt the color of the style-transferred images.
Despite the application of common image distortions, such as JPEG compression and Gaussian blur, LAACA maintains its effectiveness in both preserving color information in the style images and disrupting color in the style-transferred images, as evidenced by the consistent ACDM scores across all scenarios.
\begin{table}[t]
\caption{ACDM evaluation results, a lower score means better quality.}
\label{tab:result:acdm}
\resizebox{\columnwidth}{!}{
\begin{tabular}{cccccccc}
\toprule
(\textbf{the best},          & \multirow{2}{*}{style images$^-$} & \multicolumn{6}{c}{Neural Style Transfer Methods}    \\
\underline{the second best}) &                               & Gatys  & AdaIN  & OST    & SANet  & EFDM   & Average $^+$\\ \midrule
LAACA                        & \underline{0.0495}            & 0.1322 & 0.1913 & 0.0711 & 0.1300 & 0.1798 & \textbf{0.1409}  \\
UAP        & 0.1350                        & 0.1475 & 0.1412 & 0.1208 & 0.1406 & 0.1317 & \underline{0.1364}  \\
JPEG 75\% Comp.              & \underline{0.0492}            & 0.1244 & 0.1821 & 0.0682 & 0.1241 & 0.1694 & 0.1196  \\
Gaussian blur                & \textbf{0.0438}               & 0.0660 & 0.0839 & 0.0493 & 0.0906 & 0.0875 & 0.0755  \\ \bottomrule
\end{tabular}
}
\end{table}
\begin{table}[t]
\caption{LPIPS evaluation results, a lower score means better quality.}
\label{tab:result:lpips}
\resizebox{\columnwidth}{!}{
\begin{tabular}{cccccccc}
\toprule
(\textbf{the best},          & \multirow{2}{*}{style images$^-$} & \multicolumn{6}{c}{Neural Style Transfer Methods}    \\
\underline{the second best}) &                               & Gatys  & AdaIN  & OST    & SANet  & EFDM   & Average$^+$ \\ \midrule
LAACA                        & \textbf{0.4043}               & 0.6080 & 0.5552 & 0.4949 & 0.5545 & 0.5723 & \textbf{0.5570}  \\
UAP        & 0.5740                        & 0.5550 & 0.4239 & 0.4850 & 0.5231 & 0.4425 & 0.4859  \\
JPEG 75\% Comp.              & \textbf{0.3985}               & 0.5944 & 0.5487 & 0.4881 & 0.5492 & 0.5653 & 0.5491  \\
Gaussian blur                & \underline{0.4226}            & 0.4503 & 0.4513 & 0.4129 & 0.4615 & 0.4647 & \underline{0.4481}  \\ \bottomrule
\end{tabular}
}
\end{table}

\paragraph{LPIPS.}
The LPIPS results in Table~\ref{tab:result:lpips} demonstrate that LAACA ranks in the top tier for original/protected style images in terms of maintaining the perception, and it performs the best on disrupting the perception of post-NST images.
This indicates that LAACA is effective in preserving the perceptual quality of the style images while successfully disrupting the perceptual similarity of the NST images.
These characteristics are crucial for the adversarial attack to be less noticeable and more effective in disrupting the neural style transfer process.
When subjected to defense methods, LAACA shows competitive results.
JPEG compression performs similarly to LAACA on both style images and NST images, suggesting that LAACA maintains its effectiveness even when the images undergo compression.
Although Gaussian blur slightly degrades LAACA's performance, the performance loss remains within an acceptable range.
This indicates that our method exhibits a certain level of robustness against potential image degradation that may occur during distribution.

\subsection{Ablation Studies}

In ablation studies, in addition to SSIMc and ACDM, we also include the $L_p$ norms as evaluation metrics to quantify the pixel-level differences between the protected and original images, which is regarded as a convention in adversarial attack researches. For all experiments, we set $T=100$, ensuring convergence.

The parameter $k$ determines the separation of high-frequency and low-frequency components in the image, with a larger $k$ resulting in a wider range of frequencies being considered as high-frequency. As $k$ increases from 3 to 6, the SSIMc scores for the style images show a slight improvement, indicating that a larger $k$ may lead to a better separation of high-frequency and low-frequency regions, resulting in improved structural similarity between the pre- and post-attack style images. However, the SSIMc scores for the other NST methods decrease, suggesting that a larger $k$ may compromise their structural similarity. The ACDM scores increase with larger $k$, indicating that a larger $k$ may introduce more color distortions in the style images while affecting the color stability of other NST methods. The $\ell_2$ distance decreases slightly for the style images as $k$ increases, implying that a larger $k$ may produce lower pixel-wise differences, but it increases for the other NST methods. The $\ell_{\infty}$ scores remain relatively stable across different $k$ values for all methods, indicating that the maximum pixel-wise difference is nearly unrelated to $k$.

The parameter $\alpha$ determines the magnitude of the update in each attack iteration. As $\alpha$ increases from 1 to 8, the SSIMc scores for the style images remain relatively stable, while the scores for the other NST methods decrease, suggesting that larger step sizes may compromise their structural similarity. The ACDM scores increase with larger $\alpha$ values, indicating that a larger step size may lead to more color distortions in the style images while also affecting the color stability of other NST methods. The $\ell_2$ distance increases for all methods with higher $\alpha$ values, implying that larger step sizes may result in greater pixel-wise differences between the pre- and post-attack images. The $\ell_{\infty}$ scores remain relatively stable across different $\alpha$ values for all methods, suggesting that the maximum pixel-wise difference is not significantly influenced by $\alpha$.

The parameter $\epsilon$ defines the maximum allowed deviation from the original image in the style transfer process. As $\epsilon$ increases from 10 to 80, the SSIMc scores for the style images remain relatively high, while the scores for the other NST methods decrease significantly, indicating that a larger perturbation range leads to a substantial decrease in their structural similarity. The ACDM scores increase with larger $\epsilon$ values, suggesting that a higher perturbation range introduces more color distortions in the style images while also affecting the color stability of other NST methods. The $\ell_2$ distance increases for all methods as $\epsilon$ grows, implying that a larger perturbation range results in greater pixel-wise differences between the pre- and post-attack images. The $\ell_{\infty}$ scores increase with higher $\epsilon$ values, indicating that the maximum pixel-wise difference between pre- and post-attack images becomes larger as $\epsilon$ increases, particularly for NSTs.

\section{Conclusion, Limitations, and Future Work}
In this work, we propose the Locally Adaptive Adversarial Color Attack, a method designed to interrupt unauthorized neural style transfer use cases. Our approach significantly degrades the quality of NST outputs while introducing acceptable perturbations, which will discourage potential infringers from using the protected artwork, because of the bad NST generation.
To supplement metrics in evaluating the performance of color-mattered tasks, we introduce an IQA, ACDM, which quantifies the color distortions between pre- and post-attack style images.
The evaluation of ACDM's performance in experiments validates its effectiveness in assessing color-related tasks.
Experiments demonstrate the efficacy of our attack method in compromising the style transfer process, resulting in significant color distortions and structural dissimilarities in NST images while maintaining the acceptable visual integrity of the post-attack style images.
However, our work has some limitations. The method's runtime on common GPUs is not optimal, and the numerous hyperparameters currently provided may not be suitable for all images.
Categorizing style images into abstract and realistic paintings, our method is more effective on abstract paintings, possibly because their bolder colors and high-frequency components provide a larger manipulation space.
Future work will focus on improving computational efficiency and exploring ways to make these parameters adaptive. We aim to enhance the method's applicability across a wider range of images and hardware configurations.
Overall, our approach offers artists a potential tool to protect their intellectual property, with the promise of mitigating or curbing electronic IP infringement.
\bibliography{reference}
\end{document}